%% file: main.tex
\PassOptionsToPackage{prologue,dvipsnames}{xcolor}
\documentclass[lettersize,journal]{IEEEtran}
\usepackage{colortbl}
\usepackage{amsmath,amsfonts}
\usepackage{algorithmic}
\usepackage{algorithm}
\usepackage{array}
\usepackage[caption=false,font=normalsize,labelfont=sf,textfont=sf]{subfig}
\usepackage[table]{xcolor}
\usepackage{multirow}
\usepackage{url}
\usepackage{graphicx}
\usepackage{diagbox}
\usepackage{cite}
\usepackage {makecell}
\usepackage[pagewise]{lineno}
\usepackage[dvipsnames]{xcolor}
\usepackage[normalem]{ulem}
\usepackage{booktabs}
\usepackage[breaklinks]{hyperref}
\usepackage{pifont}
\newcommand{\settablefont}{\fontsize{8}{12}\selectfont}

\usepackage{hyperref}
\usepackage{orcidlink}
\input{packages.tex}

\title{Rebenchmarking Unsupervised Monocular\\3D Occupancy Prediction}

\author{
Zizhan~Guo\,\orcidlink{0009-0003-7360-2192},
Yi~Feng\,\orcidlink{0009-0005-4885-0850},
Mengtan~Zhang\,\orcidlink{0009-0003-3468-7680},
Haoran~Zhang\,\orcidlink{0009-0009-7578-5897},
\\
Wei~Ye\,\orcidlink{0000-0002-3784-7788},~\IEEEmembership{Member,~IEEE},
and Rui~Fan\,\orcidlink{0000-0003-2593-6596},~\IEEEmembership{Senior Member,~IEEE}
\thanks{
(\textit{Corresponding author: Rui Fan})}
}

\begin{document}

\maketitle

\begin{abstract}
Inferring the 3D structure from a single image, particularly in occluded regions, remains a fundamental yet unsolved challenge in vision-centric autonomous driving. Existing unsupervised approaches typically train a neural radiance field and treat the network outputs as occupancy probabilities during evaluation, overlooking the inconsistency between training and evaluation protocols. Moreover, the prevalent use of 2D ground truth fails to reveal the inherent ambiguity in occluded regions caused by insufficient geometric constraints. To address these issues, this paper presents a reformulated benchmark for unsupervised monocular 3D occupancy prediction. We first interpret the variables involved in the volume rendering process and identify the most physically consistent representation of the occupancy probability. Building on these analyses, we improve existing evaluation protocols by aligning the newly identified representation with voxel-wise 3D occupancy ground truth, thereby enabling unsupervised methods to be evaluated in a manner consistent with that of supervised approaches. Additionally, to impose explicit constraints in occluded regions, we introduce an occlusion-aware polarization mechanism that incorporates multi-view visual cues to enhance discrimination between occupied and free spaces in these regions. Extensive experiments demonstrate that our approach not only significantly outperforms existing unsupervised approaches but also matches the performance of supervised ones. Our source code and evaluation protocol will be made available upon publication.
\end{abstract}
\begin{IEEEkeywords}
3D occupancy prediction, benchmark, neural radiance field.
\end{IEEEkeywords}

\section{Introduction}
\label{sec:intro}

\IEEEPARstart{3}{D} occupancy prediction, which infers the volumetric structure of real-world environments, enables unified spatial representations that support various downstream tasks in autonomous systems \cite{Cao_2022_CVPR, Huang_2023_CVPR, Li_2023_CVPR}. Most existing methods~\cite{Zhang_2023_ICCV, jiang2023symphonies} rely on supervised learning with voxel-wise annotated 3D ground truth, typically generated from sparse LiDAR point clouds~\cite{tian2023occ3d, wei2023surroundocc}. Acquiring such annotations is, nevertheless, both labor-intensive and prone to inaccuracies, thereby impeding large-scale training. In contrast, unsupervised methods~\cite{wimbauer2023behind, Han_2024_CVPR, li2024know, feng2025vipocc} based on neural radiance fields (NeRFs)~\cite{mildenhall2020nerf} avoid the need for explicit supervisory signals and realize occupancy inference from a single image, demonstrating strong potential and flexibility for real-world applications.

As 3D occupancy prediction continues to advance, the systematic evaluation of emerging networks has become increasingly critical~\cite{zhang2024vision}. While supervised methods can be evaluated on well-established benchmarks~\cite{wang2025uniocc, tian2023occ3d, wei2023surroundocc, wang2023openoccupancy}, unsupervised NeRF-based approaches, developed since BTS~\cite{wimbauer2023behind}, are still evaluated using inappropriate protocols misaligned with the 3D nature of the task. Specifically, NeRF networks are designed to output implicit rendering weights for alpha compositing. As pointed out by \cite{Ahn_2024_CVPR}, the magnitude of these weights depends on the scale of the sampling interval. However, existing evaluation protocols erroneously equate these scale-variant, point-wise weights with fixed-range, voxel-wise occupancy ground truth, thereby introducing inconsistencies between the training and evaluation protocols.
In addition, existing occupancy annotations are technically limited to a 2D plane, which is ill-suited for an inherently 3D task, thereby undermining both the reliability and completeness of the evaluation results.

The aforementioned issues in current evaluation protocols obscure the inherent limitations of existing methods. Following the NeRF paradigm, early representative monocular approaches~\cite{wimbauer2023behind, Han_2024_CVPR, li2024know, feng2025vipocc} reconstruct target-view images from multiple source views through volume rendering. These networks are trained by minimizing the photometric discrepancies between reconstructed and real images. However, during the volume rendering process, density values in occluded regions contribute minimally to the reconstructed image, as image intensities from these areas are rarely transmitted through foreground occluders during rendering integration for the target view. Compared to supervised approaches, which can directly learn from occluded occupancy ground truth, NeRF-based networks inherently struggle to accurately model occupancy distributions in these regions with only 2D supervision. When the ground truth dimension is lifted to 3D, the accuracy of existing methods deteriorates greatly due to the increased proportion of occluded regions.

Therefore, in this study, we rebenchmark the unsupervised monocular 3D occupancy prediction task to address all the aforementioned challenges. First, we systematically analyze and interpret the occupancy probability in NeRF-based methods, and incorporate spatial neighborhood into point-wise occupancy estimations. This integration mitigates the magnitude variations of network outputs and alleviates spatial misalignment with voxel-wise ground truth. Furthermore, we transform the original camera coordinate system into a new space and develop an occupancy sampling algorithm to align the spatial distribution of the proposed occupancy representation with that of the 3D occupancy annotations. This algorithm enables a reliable and interpretable benchmark aligned with the standard 3D evaluation protocols widely used for supervised methods~\cite{li2024sscbench}. Moreover, we design an occlusion-aware occupancy polarization mechanism by correlating image intensity variations with occupancy discrepancies across multiple views to provide additional supervisory signals for occluded regions. Extensive experimental results on the KITTI-360~\cite{Liao2022PAMI} dataset validate both the interpretability and rationality of our reformulated benchmark, as well as the effectiveness of the proposed occupancy polarization mechanism. In addition, comprehensive comparisons with supervised methods underscore the state-of-the-art (SoTA) performance achieved by our unsupervised approach. In a nutshell, the key contributions of this study are as follows:
\begin{itemize}
\item We delve into the interpretation of occupancy probability in NeRF, bridging the gap between NeRF-based predictions and voxel-wise 3D occupancy evaluation protocols.
\item We develop a coordinated-transformed sampling algorithm that unifies the benchmark for both unsupervised and supervised 3D occupancy prediction approaches.
\item We propose an occlusion-aware occupancy polarization mechanism that exploits visual cues from other views to provide additional supervision in occluded areas.
\end{itemize}

The remainder of this article is structured as follows: Sect. \ref{sec:related} reviews existing supervised and unsupervised 3D occupancy prediction works. Sect. \ref{sec:method} presents the proposed occupancy probability interpretation, the coordinate-transformed occupancy sampling algorithm, and the proposed occlusion-aware occupancy polarization mechanism. Sect. \ref{sec:experiments} details the implementation of the proposed benchmark, compares our method with other state-of-the-art (SoTA) methods and presents the ablation study results. Limitations and corresponding analysis of our proposed method are discussed in Sect. \ref{sec:discussion}. Finally, in Sect. \ref{sec:conclusion}, we summarize this article and discuss possible directions for future work.

\section{Related Work}
\label{sec:related}

\subsection{Supervised 3D Occupancy Prediction}

Learning voxel-wise 3D occupancy from images is a key step toward comprehensive 3D scene understanding~\cite{NEURIPS2026_d8ca28a3, Ma_2024_CVPR, Li_2025_CVPR, zheng2024occworld, Wu_2025_ICCV, Zhao_2024_CVPR, Zuo_2025_CVPR, huang2024gaussianformer, Huang_2025_CVPR, Chambon_2025_ICCV}. As a pioneering study, MonoScene~\cite{Cao_2022_CVPR} introduces a 3D occupancy prediction framework that infers voxel-level geometry and semantics from a single image. TPVFormer~\cite{Huang_2023_CVPR} extends this approach to multi-camera settings by incorporating tri-perspective representations. Subsequent studies have progressively refined network architectures within this end-to-end framework. For instance, VoxFormer~\cite{Li_2023_CVPR} adopts Transformers over sparse voxels for long-range context modeling, while OccFormer~\cite{Zhang_2023_ICCV} leverages a dual-path Transformer architecture to fuse semantic and geometric features across multiple views. Building upon these prior works, Symphonies~\cite{jiang2023symphonies} leverages contextual instance queries to enhance scene-level geometric and semantic understanding in complex driving scenes. Beyond these end-to-end frameworks that directly infer voxel-level occupancies, recent methods have explored implicit representations to improve both accuracy and interpretability.
HybridOcc~\cite{zhao2024hybridocc} bridges explicit and implicit representations by integrating NeRF branches with Transformer-based voxel queries, which leads to significantly improved performance.
Building upon this line of research, the method proposed in~\cite{10379527} further advances monocular 3D occupancy prediction by extending it to the instance level through a novel network architecture design.
Nevertheless, these methods rely on the costly process of acquiring accurate 3D annotations, which limits their scalability for large-scale training.
Thus, this study focuses extensively on unsupervised methods. 

\subsection{Unsupervised 3D Occupancy Prediction}

Unsupervised methods aim to reconstruct 3D scene geometry with only 2D supervision~\cite{Jevtic_2025_SceneDINO, huang2024selfocc, zhang2025occnerf, Agro_2024_CVPR, 10947319, NEURIPS2022_19a0a55f, Jiang_2025_CVPR}.
Early unsupervised approaches~\cite{feng2025self, zhang2025dcpi, bello2024self, li2023sense, gonzalez2021plade, zhang2022self} typically relies on photometric consistency constraints between adjacent frames in monocular videos, which limits their capability to estimating per-frame monocular depth rather than recovering full 3D scene geometry.
In recent years, existing methods have been built upon NeRFs~\cite{mildenhall2020nerf}, which utilize a continuous volume rendering mechanism and optimize the network by minimizing the photometric loss across multiple views.
For multi-view unsupervised 3D occupancy prediction tasks, RenderOcc~\cite{10611537} is presented by extracting a NeRF-style 3D volume representation from multi-view images and establishing 2D volume renderings, thus enabling direct 3D supervision from 2D semantics and depth labels.
BTS~\cite{wimbauer2023behind} is a pioneering work that presents a fully unsupervised NeRF pipeline for single-view 3D reconstruction through differentiable volume rendering. Building upon this foundation, KDBTS~\cite{Han_2024_CVPR} distills multi-view density fields into a single-view network via self-supervised training, thereby greatly improving its performance across diverse scenes. Subsequent studies~\cite{li2024know, feng2025vipocc} have increasingly incorporated off-the-shelf vision models to improve object-level 3D occupancy predictions. For instance, KYN~\cite{li2024know} leverages vision-language priors to integrate semantic knowledge and spatial context into the pipeline for semantically guided 3D geometric reasoning. ViPOcc~\cite{feng2025vipocc} further introduces visual priors from foundation models to enhance instance-level semantic reasoning and temporal photometric consistency. Despite these advances, existing approaches remain constrained by their reliance primarily on a reconstruction loss through volume rendering, which inherently fails to provide explicit guidance in occluded regions. Additionally, they often overlook the inconsistency between training and evaluation protocols, ultimately compromising the reliability of 3D occupancy predictions. In this work, we present a more interpretable representation of occupancy probability and propose an occlusion-aware polarization mechanism to solve these issues.

\subsection{Benchmarks for 3D Occupancy Prediction}

Several datasets~\cite{Liao2022PAMI, caesar2020nuscenes} provide video sequences accompanied by camera poses and LiDAR point clouds collected in real-world driving environments. To enable 3D occupancy prediction evaluation, recent studies~\cite{wei2023surroundocc, li2024sscbench} have constructed voxel-level occupancy annotations by aggregating multi-frame LiDAR point clouds. Occ3D~\cite{tian2023occ3d} is among the first to achieve voxel-level semantic annotations, enabling dense 3D occupancy evaluation at fine granularity. SurroundOcc~\cite{wei2023surroundocc} applies Poisson reconstruction to consolidate LiDAR scans into dense 3D annotations, while OpenOccupancy~\cite{wang2023openoccupancy} improves labeling accuracy through extensive manual annotation to mitigate LiDAR sparsity. Recent benchmarks such as SSCBench~\cite{li2024sscbench} and UniOcc~\cite{wang2025uniocc} extend unified evaluation protocols to a variety of driving scenes. Despite recent progress focused primarily on supporting supervised learning paradigms, a standardized benchmark for unsupervised 3D occupancy learning remains underdeveloped, with existing annotations often limited to a single 2D plane. In this work, we exclusively utilize the aforementioned voxel-level occupancy annotations to evaluate unsupervised approaches, thereby establishing a comprehensive 3D benchmarking protocol.

\section{Methodology}
\label{sec:method}

\subsection{Problem Setup}
In NeRF-based approaches, the network with parameters $\boldsymbol{\Theta}$ takes as input
a target-view RGB image $\boldsymbol{I}_{0}$, camera intrinsic matrix $\boldsymbol{K}$, and a 3D point $\boldsymbol{x}^{(i)}$ to predict a rendering weight $\sigma ^{(i)}$ as follows:
\begin{equation}
  \sigma^{(i)} = f\left(\boldsymbol{I}_{0}, \boldsymbol{K}, \boldsymbol{x}^{(i)}; \boldsymbol{\Theta}\right), 
  \label{eq:infer}
\end{equation}
where the 3D point $\boldsymbol{x}^{(i)} = \boldsymbol{o} + t^{(i)} \boldsymbol{d}$ is sampled along a ray in the direction of the unit vector $\boldsymbol{d}$, with $t^{(i)}$ denoting the distance from the camera origin $\boldsymbol{o}$ to $\boldsymbol{x}^{(i)}$.
During the volume rendering process, the opacity $\alpha^{(i)}$ is first computed along the ray using the following expression:
\begin{equation}
\alpha^{(i)} = 1 - \exp \left(- \sigma^{(i)} \delta^{(i)} \right),
\label{eq:alpha}
\end{equation}
where $\delta^{(i)} = |\boldsymbol{x}^{(i+1)} - \boldsymbol{x}^{(i)}|$ denotes the length of the ray segment between consecutive sample points $\boldsymbol{x} ^{(i)}$ and $\boldsymbol{x}^{(i+1)}$.
An image in the target view is rendered as follows:
\begin{equation}
  \hat{\boldsymbol{c}} = \sum_{i=1}^{N} \alpha^{(i)} T^{(i)} \boldsymbol{c}^{(i)}, \quad
  T ^{(i)} = \prod _{j=1}^{i-1} \left(1 - \alpha ^{(j)}\right),
  \label{eq:render}
\end{equation}
where $\hat{\boldsymbol{c}}$ denotes the image intensity rendered along a sampled ray with $N$ sampled points, $\boldsymbol{c}^{(i)}$ denotes the color at point $\boldsymbol{x}^{(i)}$ sampled from other viewpoints, and $T^{(i)}$ represents the accumulated transmittance up to the $i$-th point.
The network is trained by minimizing a photometric loss objective that quantifies the discrepancy between the rendered and ground-truth RGB intensities.

\subsection{Occupancy Probability Interpretation for NeRF}
\label{sec:opi}

\begin{figure}[t!]
    \centering
    \includegraphics[width=0.99\linewidth]{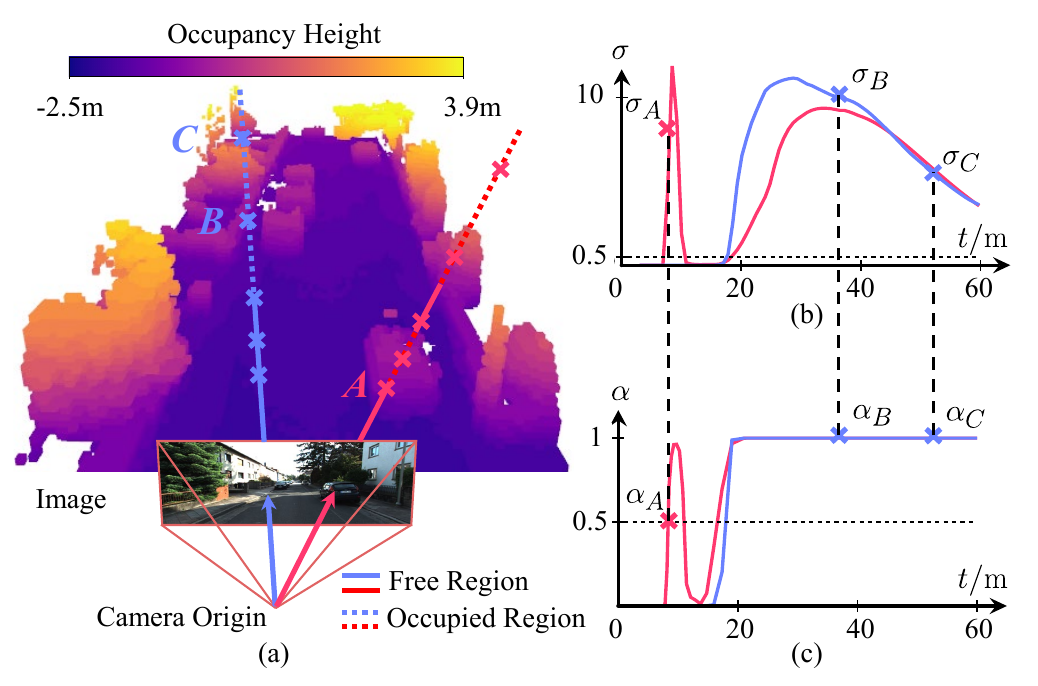}
    \caption{
        A comparison between the network output $\sigma$ and the opacity $\alpha$ during inference: (a) two representative sampled rays; (b) $\sigma$ distributions; (c) $\alpha$ distributions. 
        For point A, which transitions from occupied to free space, $\alpha_A$ is bounded within the range $(0, 1)$, whereas $\sigma_A$ has no upper bound, making $\alpha$ a more suitable representation for occupancy probability;
        For points B and C with identical occupancy status, their discrepancy in $\sigma$ is significantly greater than that in $\alpha$, demonstrating that our proposed representation for occupancy probability effectively eliminates the magnitude variation caused by non-uniform point sampling.}
    \label{fig:distribution-interpretation}
\end{figure}

Existing evaluation protocols simply adopt a fixed $\sigma$ threshold of 0.5 to binarize each voxel\footnote{In this subsection, we omit the superscript $^{(i)}$ for $\sigma$, $\alpha$, $\delta$, and $T$ for notational simplicity.}:
\begin{equation}
o (\boldsymbol{x}) = [\sigma > 0.5],
\label{eq:old}
\end{equation}
where $[\cdot]$ denotes the Iverson bracket, and $\boldsymbol{x}$ represents the voxel center.
However, according to Eq.~\ref{eq:infer}, the network predicts occupancy densities at infinitesimal points, whereas the ground truth typically corresponds to the occupied probability of a volumetric cell. The spatial misalignment between point-wise predictions and voxel-level ground truth annotations makes their direct comparison with a fixed threshold of 0.5 in Eq.~\ref{eq:old} uninterpretable.

Specifically, existing approaches typically perform non-uniform point sampling along each ray during training, with denser sampling near the camera and sparser sampling in more distant regions. The variation in sampling density along the ray changes the spatial neighborhood around each point, which is characterized by the distance $\delta$ to its nearest point. Consider two occupied points, B and C, as illustrated in Fig.~\ref{fig:distribution-interpretation}(a), with the former located in a densely sampled region and the latter located in a sparsely sampled one. According to Eq.~\ref{eq:render}, the rendering contributions $\alpha T$ of these two occupied points to the reconstructed pixel color are theoretically expected to be equivalent. In the training process, the network adjusts its output $\sigma$ to satisfy this equality, which leads to a magnitude variation issue, as illustrated in Fig.~\ref{fig:distribution-interpretation}(b). As indicated by Eq.~\ref{eq:alpha}, denser sampling yields a smaller spatial neighborhood $\delta_B$, which in turn increases the network output $\sigma_B$, and conversely, a larger $\delta_C$ in sparse sampling regions results in a lower $\sigma_{C}$. This dependency established during training is retained at inference time, causing magnitude variations of the network outputs along the ray. Current method, nevertheless, overlooks this variation and applies a fixed threshold to obtain voxel-wise occupancy predictions, leading to inconsistencies in existing evaluation protocols across regions with varying point sampling densities. In addition, when the number of sampled points along each ray changes, the spatial neighborhood $\delta$ around each point also varies, thereby inducing inconsistent output magnitudes due to the underlying dependency.
This also leads to inconsistency of evaluation protocols across different experimental configurations.

In this study, we argue that the opacity $\alpha$ provides a more interpretable representation of occupancy probability than the network's output $\sigma$. According to Eq.~\ref{eq:alpha}, the opacity, computed based on the spatial neighborhood $\delta$, characterizes the occupancy attributes within a finite volume rather than at an infinitesimal point. This volumetric interpretation aligns more naturally with voxel-level occupancy ground truth, making it better suited for model evaluation. In addition, unlike the original network output $\sigma$, which is highly sensitive to the point sampling strategy, the opacity value is bounded within $(0, 1)$, as illustrated in Fig.~\ref{fig:distribution-interpretation}(c). This bounded range eliminates the aforementioned magnitude variation effect, thereby enhancing the consistency of the evaluation protocols across diverse settings. The occupancy prediction is formulated under the proposed interpretation as follows:
\begin{equation}
    o(\boldsymbol{x}) = [ \alpha > 0.5].
    \label{eq:new_simplified}
\end{equation}
By adopting this opacity-based volumetric interpretation, we redefine the occupancy probability representation and reformulate the entire benchmark.

\subsection{Coordinate-Transformed Occupancy Sampling}

\begin{figure}[t!]
    \centering
    \includegraphics[width=0.99\linewidth]{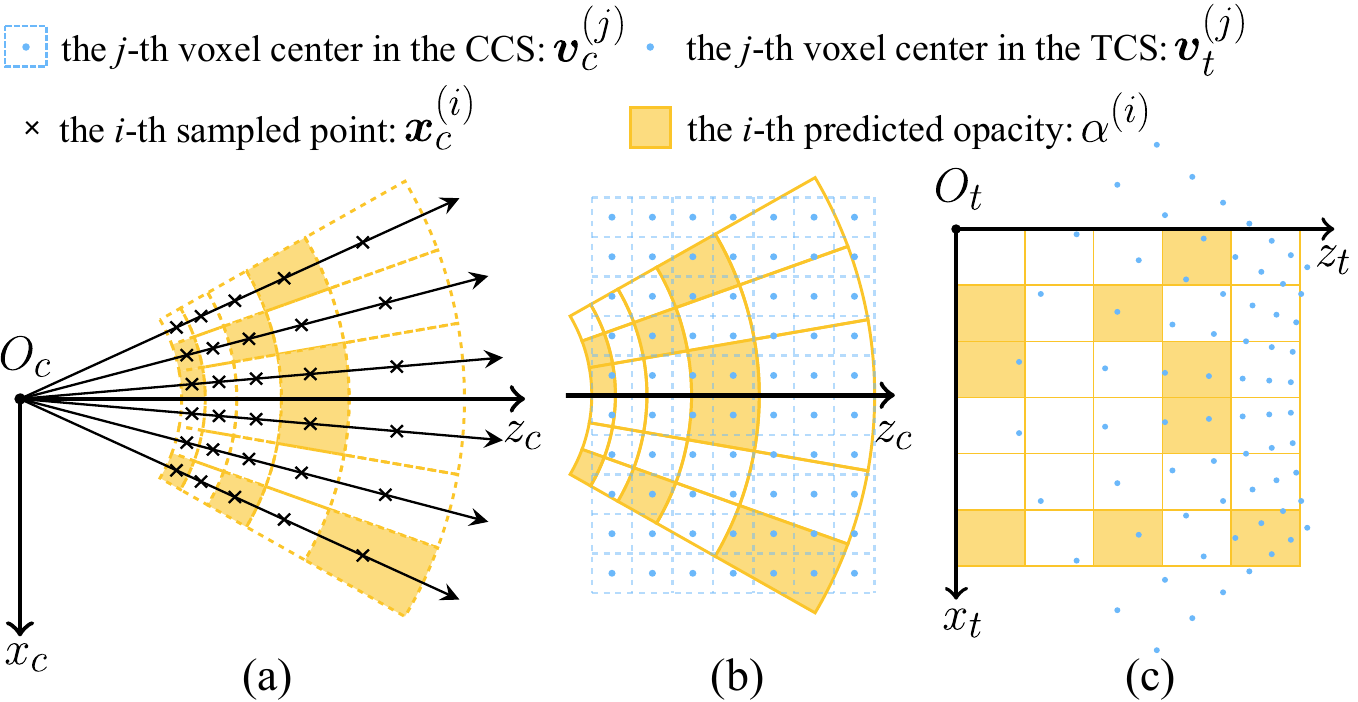}
    \caption{
        The occupancy sampling algorithm in the camera coordinate system (CCS) and the transformed coordinate system (TCS): (a) network inference with sampled points as input; (b) opacity distribution v.s. the voxel grid in the CCS; (c) opacity sampling using voxel centers in the TCS.
      }
    \label{fig:voxelization}
\end{figure}

As discussed in the previous subsection, opacity $\alpha$ provides a more appropriate representation for occupancy prediction. Nonetheless, as illustrated in Figs.~\ref{fig:voxelization}(a) and (b), the predicted opacities are distributed along radial segments originating from the camera center, whereas the ground-truth occupancy annotations are defined on a uniform voxel grid. To address this spatial misalignment problem, we propose a coordinate-transformed occupancy sampling algorithm that maps opacities from radial segments onto the voxel grid.

As illustrated in Fig.~\ref{fig:voxelization}(c), to more clearly characterize the radial distribution of opacity, we construct the TCS, in which opacity is uniformly distributed, from the CCS. A 3D point $\boldsymbol{x} _{c} = (x_c, y_c, z_c)^{\top}$ in the CCS corresponds to the homogeneous pixel coordinates $\begin{pmatrix} u, v, 1 \end{pmatrix}^{\top} = \boldsymbol{K} \boldsymbol{x}_c / z_c$, where $\boldsymbol{K}$ denotes the camera intrinsic matrix. The coordinates of the corresponding point in the TCS can be computed using the following expression:
\begin{equation}
    \boldsymbol{x}_{t} = 
    \begin{pmatrix}
        \dfrac{u}{w-1}, \dfrac{v}{h-1}, \dfrac{1 / t_n - 1/\Vert\boldsymbol{x}_c\Vert _{2}}{1/t_n - 1/t_f}
    \end{pmatrix}^{\top},
    \label{eq:cTp}
\end{equation}
where the image resolution is $h \times w$ pixels, and $t_{n}$ and $t_{f}$ denote the near and far bounds, respectively. This transformation maps the view frustum in the rendering field into a normalized cube spanning $[0, 1]^3$, where the $x$- and $y$-axes in the TCS are aligned with the image axes, and the $z$-axis is aligned with the sampled ray direction. The TCS enables the grid sampling process, which requires a uniformly distributed opacity map.

Given the above details on the defined TCS, the coordinate-transformed occupancy sampling algorithm proceeds with the following steps. First, each 3D point $\boldsymbol{x}^{(i)}_c$ is sampled in the CCS using the same strategy adopted during training. Following Eq.~\ref{eq:infer}, $\boldsymbol{x}^{(i)}_c$ is passed through the network to infer the cooresponding output $\sigma^{(i)}$, as depicted in Fig.~\ref{fig:voxelization}(a). Following Eq.~\ref{eq:alpha}, the opacity $\alpha^{(i)}$ is then computed along each camera ray using the sampling interval $\delta^{(i)}$, as shown in Fig.~\ref{fig:voxelization}(b).
Subsequently, each voxel center $\boldsymbol{v}_c^{(j)}$ is transformed from the CCS to the TCS, yielding the corresponding coordinates $\boldsymbol{v}_t^{(j)}$, as illustrated in Fig.~\ref{fig:voxelization}(c). The voxel-wise occupancy predictions, spatially aligned with the ground truth annotations, are obtained by sampling the grid-based opacity map in the TCS, expressed as follows:
\begin{equation}
  o (\boldsymbol{v} _{c}^{(j)}) = \left[\mathcal{A} \langle \boldsymbol{v} _{t} ^{(j)} \rangle > 0.5\right],
  \label{eq:new}
\end{equation}
where $\mathcal{A}\langle \cdot \rangle$ denotes the grid sampling process on the opacity map $\mathcal{A} = \{ \alpha^{(i)} \}$. The resulting occupancy predictions are subsequently evaluated using the metrics in our benchmark.

\subsection{Occlusion-Aware Occupancy Polarization}
\label{sec:op}

Further exploration of our benchmark reveals a critical limitation of existing unsupervised NeRF-based occupancy prediction methods: they inherently struggle to predict occupancy distributions behind foreground occluders. This limitation arises from their exclusive reliance on reconstruction loss derived from volume rendering. To better understand this limitation, we conduct a quantitative analysis on the supervisory signals within occluded regions during the back-propagation process.

For simplicity, we define the per-pixel photometric reconstruction loss as $\mathcal{L}_{r} = \left| \hat{\boldsymbol{c}} - \boldsymbol{c}_{gt} \right|$, where $\boldsymbol{c}_{gt}$ denotes the ground-truth RGB value.
We derive the gradient of this reconstruction loss with respect to the opacity at the sampled point $\boldsymbol{x}^{(i)}$ as follows:
\begin{equation}
    \frac{\partial \mathcal{L}_{r}}{\partial \alpha^{(i)}} = \frac{\partial \mathcal{L}_{r}}{\partial \hat{\boldsymbol{c}}} \frac{\partial \hat{\boldsymbol{c}}}{\partial \alpha^{(i)}}
    = [ \hat{\boldsymbol{c}} > \boldsymbol{c}_{gt} ]^{\top} \frac{\partial \hat{\boldsymbol{c}}}{\partial \alpha^{(i)}},
\label{eq:dLdc}
\end{equation}
where $[\cdot]$ denotes the Iverson bracket.
We take the derivate of the rendered pixel color $\hat{\boldsymbol{c}}$ and obtain the following expression:
\begin{equation}
  \begin{split}
    \frac{\partial \hat{\boldsymbol{c}}}{\partial \alpha^{(i)}}
    &= \sum\limits_{j=1}^{N} \frac{\partial}{\partial \alpha^{(i)}}
    \left(
      \alpha^{(j)} T^{(j)} \boldsymbol{c}^{(j)
      }
    \right) \\
    &= T^{(i)} \boldsymbol{c}^{(i)}
    + \sum\limits_{j=1}^{N} \alpha^{(j)} \boldsymbol{c}^{(j)} \frac{\partial T^{(j)}}{\partial \alpha^{(i)}},
  \end{split}
  \label{eq:derivate}
\end{equation}
where ${\partial T^{(j)}}/{\partial \alpha^{(i)}}$ is obtained by taking the derivative of $T^{(j)} = \prod\limits_{k=1}^{j-1} \left( 1 - \alpha^{(k)}\right)$. We present the derivative by cases as follows:
\begin{equation}
  \frac{\partial T^{(j)}}{\partial \alpha^{(i)}} =
  \left\{
    \begin{aligned}
      &0 && (j \leq i) \\
      &-T^{(i-1)} \prod\limits_{k=i+1}^{j-1}(1 - \alpha^{(k)}) && (j > i)
    \end{aligned}
  \right..
  \label{eq:dTda}
\end{equation}
We combine the above expressions and yield the final derivative, expressed as follows:
\begin{equation}
\begin{aligned}
\frac{\partial \mathcal{L}_{r}}{\partial \alpha^{(i)}}
= {} & [\hat{\boldsymbol{c}} > \boldsymbol{c}_{gt}]^{\top}
\Bigg(
    T^{(i)} \boldsymbol{c}^{(i)} \\
&   - T^{(i-1)} \sum\limits_{j=i+1}^{N}
      \alpha^{(i)} \prod\limits_{k=i+1}^{j-1}
      (1 - \alpha^{(k)}) \boldsymbol{c}^{(j)}
\Bigg).
\end{aligned}
\label{eq:back-propagation}
\end{equation}

As shown in Eq.~\ref{eq:render}, the transmittance $T^{(i)}$ decreases monotonically as the depth of $\boldsymbol{x}^{(i)}$ increases, approaching zero in regions occluded by foreground occluders. According to Eq.~\ref{eq:back-propagation}, when both $T^{(i-1)}$ and $T^{(i)}$ approach zero, the gradient ${\partial L_{r}}/{\partial \alpha^{(i)}}$ diminishes as well. As a result, the gradients with respect to the network parameters in occluded regions become negligible, thereby hindering effective learning in these areas.

\begin{figure}[t!]
    \centering
    \includegraphics[width=0.99\linewidth]{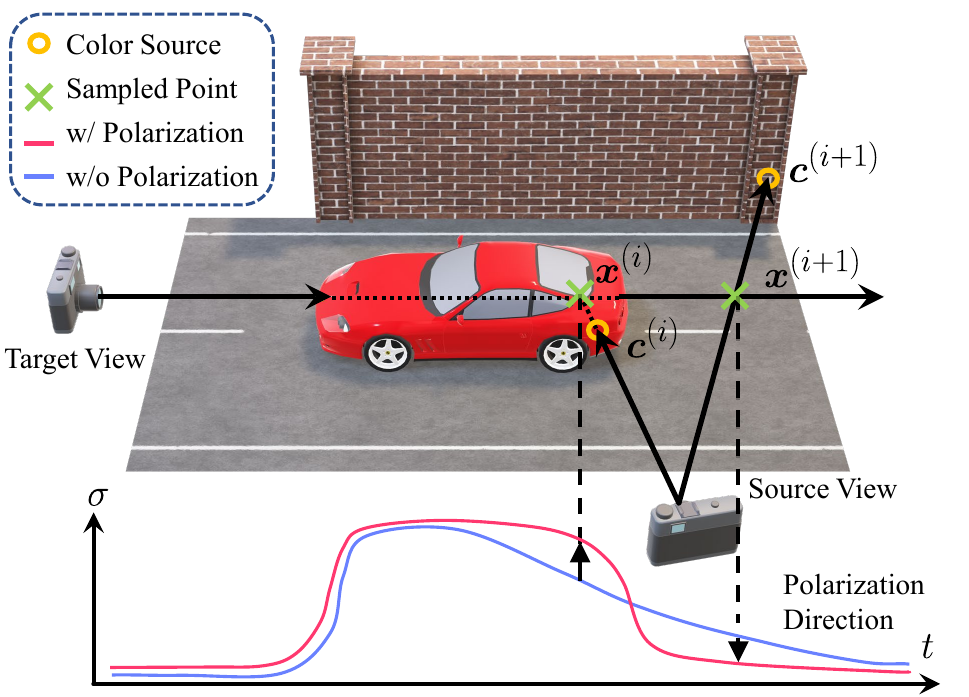}
    \caption{
        An illustration of the occlusion-aware occupancy polarization mechanism.
        For adjacent occluded points in the target view, the discrepancy in their sampled colors from the source view indicates that the colors likely originate from distinct objects.
        The proposed mechanism amplifies the occupancy differences between such points, enabling the network to refine predictions in occluded regions.
    }
    \label{fig:occlusion-aware-constraint}
\end{figure}

To address this issue, we leverage visual cues from other views to incorporate additional, explicit supervisory signals. As illustrated in Fig.~\ref{fig:occlusion-aware-constraint}, although occupancy in occluded regions is invisible in the target view, it may become visible in certain source views. Consider two adjacent sampled points, $\boldsymbol{x}^{(i)}$ and $\boldsymbol{x}^{(i+1)}$, located along a single ray. If at least one of them, \textit{e.g.}, $\boldsymbol{x}^{(i)}$, is occupied, the color difference between the two points provides valuable visual cues about the underlying occupancy. When the neighboring point $\boldsymbol{x}^{(i+1)}$ is unoccupied, its sampled color $\boldsymbol{c}^{(i+1)}$ often originates from a surface different from that of $\boldsymbol{x}^{(i)}$, leading to a noticeable color discrepancy. In contrast, if both points are occupied and lie on the same object, their colors tend to be similar due to their shared surface. Thus, the color discrepancy or similarity between adjacent points serves as an effective indicator of local occupancy variation. Nonetheless, when both points are unoccupied, the observed color difference is typically unrelated to underlying scene geometry and provides limited value for occupancy learning.

Motivated by the observation, we develop an occlusion-aware occupancy polarization mechanism to explicitly guide occupancy predictions across occupied and free space, enhancing the supervision signals in occluded regions.
Specifically, this mechanism encourages the network to polarize the predicted occupancy values of adjacent points $\boldsymbol{x}^{(i)}$ and $\boldsymbol{x}^{(i+1)}$ when their sampled colors differ significantly. This facilitates sharper occupancy transitions in regions where visual cues suggest a boundary. We implement this mechanism by formulating a polarization loss $\mathcal{L}_p$ as follows:
\begin{equation}
    \mathcal{L}_{p} = \sum\limits_{i=1}^{N-1} M_{i} \left| \boldsymbol{c}^{(i+1)} - \boldsymbol{c}^{(i)} \right|
    \text{exp}\left({- \left|\sigma^{(i+1)} - \sigma^{(i)} \right| }\right),
    \label{eq:polar}
\end{equation}
where $M_{i} = \mathrm{max}(\alpha^{(i)}, \alpha^{(i+1)})$ is a weighting mask to exclude regions where both consecutive points are unoccupied, as discussed above. The loss penalizes insufficient polarization across object boundaries and diminishes as the predicted occupancy difference increases.
The overall loss is:
\begin{equation}
    \mathcal{L} = \lambda_{r} \mathcal{L}_{r} + \lambda_{p} \mathcal{L}_{p},
\end{equation}
where $\lambda_{r}$ and $\lambda_{p}$ are the weighting parameters.

\section{Experiments}
\label{sec:experiments}

\subsection{Benchmark}

Existing evaluation protocols for unsupervised approaches are typically restricted to a narrow 2D slice of the scene, with highly limited spatial ranges ($y=0.375\mathrm{m}$, $x\in[-4 \mathrm{m}, +4 \mathrm{m}]$, and $z\in[3 \mathrm{m}, 20 \mathrm{m}]$ in the CCS).
To overcome this limitation, we utilize the 3D occupancy ground truth from SSCBench-KITTI-360 dataset~\cite{li2024sscbench}, which covers a substantially larger spatial volume extending $51.2\mathrm{m}$ forward, $25.6\mathrm{m}$ to each side, and $6.4\mathrm{m}$ in height, discretized into a $256 \times 256 \times 32$ voxel grid with a resolution of $0.2\mathrm{m}$.
Specifically, we align the 3D occupancy annotations for each image in the KITTI-360 test set and provide the transformation from the voxel coordinate system to the camera coordinate system. With this transformation, we generate 3D frustum masks and 3D voxel visibility masks using a ray-tracing algorithm, thereby enabling evaluation in occluded regions.
During inference, we apply the coordinate-transformed occupancy sampling algorithm to map the predicted opacities into the voxel grid, facilitating quantitative evaluation with voxel-based occupancy ground truth.

\subsubsection{Dataset Temporal Alignment}
The SSCBench-KITTI-360 dataset~\cite{li2024sscbench} provides 3D occupancy ground truth along with the corresponding 2D images. 
However, it employs a non-public method to select 2D image frames from the KITTI-360 dataset, resulting in updated image indices. As a result, the provided occupancy ground truth cannot be directly aligned with the test images from the KITTI-360 dataset. To address this issue, we systematically scan both datasets and construct a frame correspondence lookup table by identifying exact matches between 2D images.

\subsubsection{Transformation between Coordinate Systems}
In the SSCBench-KITTI-360 dataset, ground-truth 3D occupancy annotations are provided at a lower temporal resolution to reduce storage cost. Specifically, for every five consecutively reindexed frames, only one frame is associated with a 3D occupancy label.
Fortunately, based on our experimental verification, all selected frames are accompanied by ground-truth poses from the original KITTI-360 dataset. This enables us to compute the relative pose transformation between a query frame in the test split and its nearest adjacent frame with ground-truth 3D occupancy annotations. Consequently, we can derive the transformation from the voxel coordinate system of the annotated occupancy grid to the camera coordinate system of the test frame.

\begin{figure}[t!]
    \centering
    \includegraphics[width=0.99\linewidth]{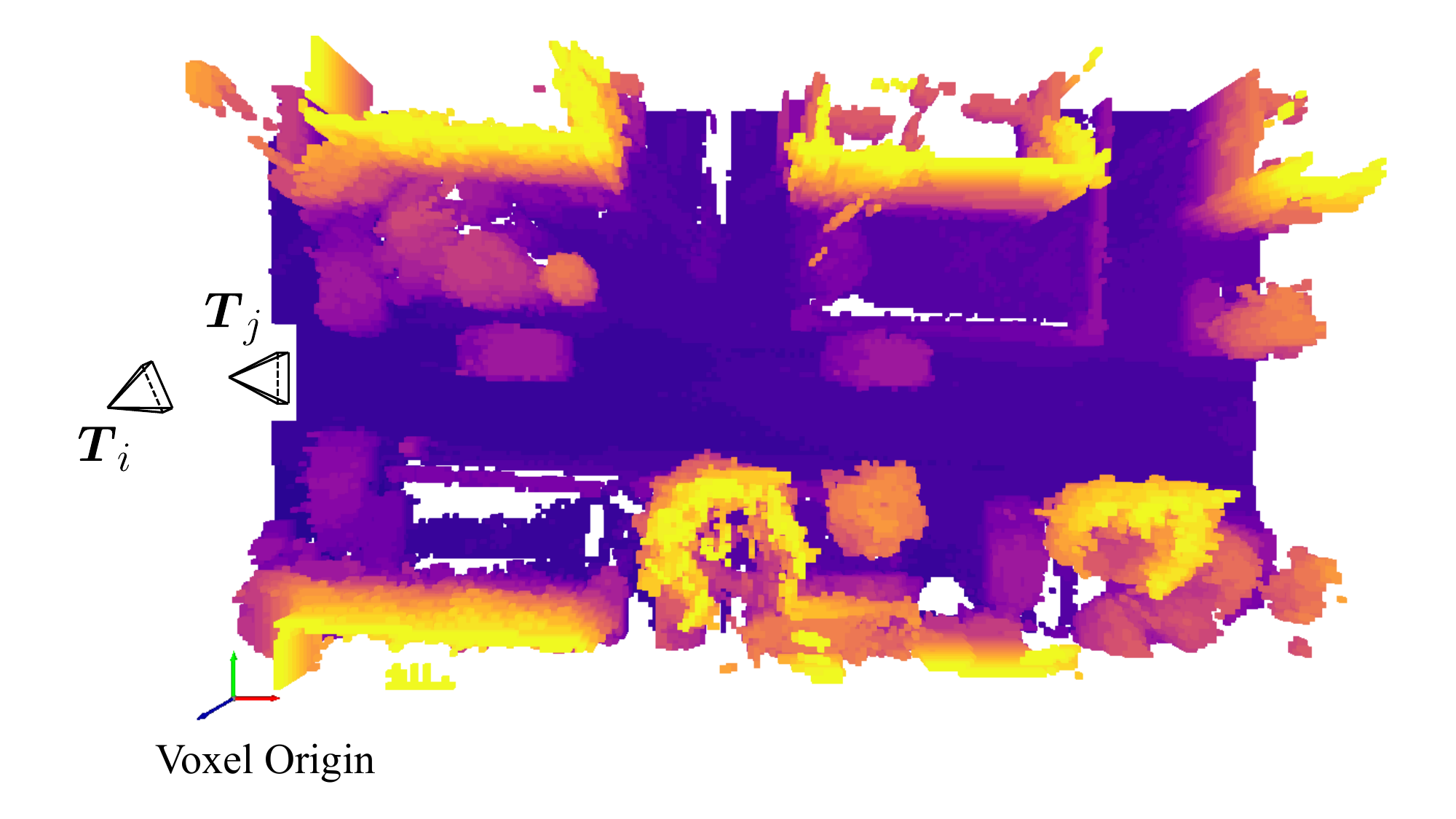}
    \caption{
      An illustration of the coordinate system transformation is provided in the bird’s-eye view of the occupancy ground truth.
      Specifically, it depicts the transformation from the voxel coordinate system associated with the $j$-th frame to the camera coordinate system of the $i$-th frame.
      This transformation enables subsequent computation of the frustum mask and the visibility mask.
    }
    \label{fig:transformation}
\end{figure}

Specifically, for the $i$-th index in the test split, we first identify the nearest subsequent index $j$ such that the corresponding frame has an associated 3D occupancy annotation in the SSCBench-KITTI-360 dataset, as illustrated in Fig.~\ref{fig:transformation}.
For example, the first test frame (index: 386) in the KITTI-360 dataset lacks an associated 3D annotation, but its nearest subsequent frame (index: 387) has a 3D occupancy label, which corresponds to index 295 in the SSCBench-KITTI-360 dataset.
This mapping ensures evaluation even when direct 3D annotations are unavailable for specific test frames. Let $\boldsymbol{T}_i$ and $\boldsymbol{T}_j$ denote the ego poses at indices $i$ and $j$, respectively. The transformation from the LiDAR coordiante system to the camera coordinate system provided by the KITTI-360 dataset is denoted by $\boldsymbol{T}_{l \to c}$, while the transformation from the voxel coordinate system to the LiDAR coordinate system provided by the SSCBench-KITTI-360 dataset is denoted by $\boldsymbol{T}_{v \to l}$. $\boldsymbol{T}_{v \to c}$, the transformation from the voxel coordinate system to the camera coordinate system can be obtained using the following expression:
\begin{equation}
    \boldsymbol{T}_{v \to c} = \boldsymbol{T}_{i}^{-1} \boldsymbol{T}_{j} \boldsymbol{T}_{l \to c} \boldsymbol{T}_{v \to l}.
    \label{eq:transformation}
\end{equation}

\subsubsection{Mask Generation}
Based on the transformation defined in Eq.~\ref{eq:transformation}, we construct the frustum mask $\boldsymbol{M}_f$ and the visibility mask $\boldsymbol{M}_v$ corresponding to the 3D occupancy ground truth. For each voxel center $\boldsymbol{v}^{(i)}_c$, we define binary values $m^{(i)}_f$ and $m^{(i)}_v$ in this voxel to represent its frustum and visibility status, respectively. Given the projected image coordinates $(u, v)$ that correspond to the voxel center $\boldsymbol{v}_{c}^{(i)}$, the value in the frustum mask is defined as follows:
\begin{equation}
    m_f^{(i)} = [0 \leq u \leq w - 1] \land [0 \leq v \leq h - 1],
\end{equation}
where $w$ and $h$ denote the image width and height, respectively. In this equation, $m_f^{(i)} = 1$ indicates that the voxel projects within the valid image bounds.

We apply a ray tracing algorithm to generate visibility masks based on 3D occupancy ground truth. Specifically, we generate a ray for each image pixel and perform dense sampling along the ray within the spatial bounds of the 3D occupancy volume. The sampling interval is set equal to the voxel size to ensure that any two adjacent sampled points lie either within the same voxel or in adjacent voxels. Subsequently, we query the ground-truth occupancy at each sampled location to determine whether it is occupied.
For a given ray, the visibility status $m_v^{(i)}$ of a point is set to 1 if the point itself and all preceding points along the ray are unoccupied. These per-point visibility statuses are then mapped back to their corresponding voxels, resulting in the final voxel-wise visibility mask $\boldsymbol{M}_v$.

\subsubsection{Coordinate-Transformed Sampling}
In the coordinate-transformed occupancy sampling process, we transform both the opacity values inferred by the network and the voxel center points from the CCS to the TCS. This transformation enables us to compute the occupancy probability for each voxel using the grid sampling algorithm. Specifically, we perform 3D grid sampling in the normalized spanning cube $[0, 1]^3$ using the bilinear interpolation mode and the border padding mode.

\subsubsection{Occupancy Metrics}

Based on the above masks, we extend the evaluation metrics used in \cite{wimbauer2023behind} to the 3D domain. These metrics include: occupancy accuracy $\mathrm{O}_\text{Acc}$, occupancy precision $\mathrm{O}_\text{Pre}$, occupancy recall $\mathrm{O}_\text{Rec}$, invisible and empty accuracy $\mathrm{IE}_\text{Acc}$, invisible and empty precision $\mathrm{IE}_\text{Pre}$, and invisible and empty recall $\mathrm{IE}_\text{Rec}$. The first three metrics are computed within the camera frustum, while the latter three are evaluated within the intersection of the camera frustum and the invisibility mask. The expressions of these evaluation metrics is given as follows:
\begin{equation}
    \mathrm{O}_{\mathrm{Acc}} = \frac{\sum\limits_{i} \left[\hat{o}^{(i)} = o^{(i)}\right] m_f^{(i)}}{\sum\limits_{i} m_f^{(i)}},
\end{equation}
\begin{equation}
    \mathrm{O}_{\mathrm{Pre}} = \frac{\sum\limits_{i} o^{(i)}  \hat{o}^{(i)} m_f^{(i)}}{\sum\limits_{i} \hat{o}^{(i)}  m_f^{(i)}},
\end{equation}
\begin{equation}
    \mathrm{O}_{\mathrm{Rec}} = \frac{\sum\limits_{i} \hat{o}^{(i)}  o^{(i)} m_f^{(i)}}{\sum\limits_{i} o^{(i)}  m_f^{(i)}},
\end{equation}
\begin{equation}
    \mathrm{IE}_{\mathrm{Acc}} = \frac{\sum\limits_{i}\left[\hat{o}^{(i)} = o^{(i)}\right] (1 - m_v^{(i)}) m_f^{(i)}}{\sum\limits_{i} (1 - m_v^{(i)})  m_f^{(i)}},
\end{equation}
\begin{equation}
    \mathrm{IE}_{\mathrm{Pre}} = \frac{\sum\limits_{i} (1 - o^{(i)})  (1 - \hat{o}^{(i)})  (1 - m_v^{(i)})  m_f^{(i)}}{\sum\limits_{i} (1 - \hat{o}^{(i)})  (1 - m_v^{(i)})  m_f^{(i)}},
\end{equation}
\begin{equation}
    \mathrm{IE}_{\mathrm{Rec}} = \frac{\sum\limits_{i} (1 - \hat{o}^{(i)})  (1 - o^{(i)})  (1 - m_v^{(i)})  m_f^{(i)}}{\sum\limits_{i} (1 - o^{(i)})  (1 - m_v^{(i)})  m_f^{(i)}},
\end{equation}
where $\hat{o}^{(i)}$ and $o^{(i)}$ denote the predicted and ground-truth occupancies for the $i$-th voxel, respectively. In addition, to facilitate comparison with supervised methods, we adopt unified metrics, including intersection over union (IoU), precision (Pre), and recall (Rec).

\subsection{Training Details}
\subsubsection{Sampling Strategy}
To support the discussion in \ref{sec:opi} regarding the magnitude variation of the network's output induced by different sampling strategies, we provide a detailed description of the sampling strategies adopted in existing methods~\cite{wimbauer2023behind, Han_2024_CVPR, li2024know, feng2025vipocc} and ours.
These strategies are categorized as ray sampling and point sampling.

The ray sampling strategy differs between the training and evaluation phases to balance computational efficiency and performance.
During training, to reduce computational cost and accelerate network convergence, we adopt a patch sampling strategy. Specifically, we extract 64 image patches (resolution: $8 \times 8$ pixels), resulting in 4,096 sampled pixels and their corresponding rays per training iteration.
This sampling strategy ensures the spatial diversity and similarity among sampled rays, thereby facilitating network training.
During evaluation, we sample all pixels in the input images to predict 3D occupancies across the entire spatial space in the corresponding camera frustum.

After sampling rays from the input image, 3D points are sampled along each ray for network inference.
Owing to the varying depth sensitivity of pinhole cameras, existing strategies do not sample points uniformly in the Euclidean depth space.
Instead, they apply uniform sampling in the inverse-depth space, which allows for denser sampling near the camera and sparser sampling in distant areas.
The distance between the camera origin and the $i$-th sampled point on a sampled ray is expressed as follows:
\begin{equation}
  t^{(i)} = 1 /
  \left(
  \left(1 - \frac{i + r}{N}\right)\frac{1}{t_n} + \frac{i + r}{N} \frac{1}{t_f}
  \right),
  \label{eq:pd}
\end{equation}
where $N$ denotes the number of points sampled per ray, $r \sim N(-0.5, 0.5)$ denotes a random variable from a uniform distribution, and $t_n$ and $t_f$ represent the near and far bounds in the rendering field, respectively.
The sampled points are defined as $\boldsymbol{x}_c^{(i)} = \boldsymbol{o} + t^{(i)} \boldsymbol{d}$, where $\boldsymbol{o}$ denotes the camera origin and $\boldsymbol{d}$ represents the unit direction vector of the sampled ray.
As discussed in \ref{sec:opi}, this non-uniform sampling strategy introduces variations in the network output magnitudes due to the depth-dependent sampling density of the points.
During evaluation, instead of directly using voxel centers as input points to the network, we employ the same point sampling strategy as used during training, with the only difference being the removal of the random variable $r$.
This eliminates the inconsistency in the spatial distribution of sampled points across both phases.

\subsubsection{Datasets Details}
We conduct extensive experiments on the KITTI-360 dataset~\cite{Liao2022PAMI}, with 3D occupancy ground truth provided by the SSCBench-KITTI-360 dataset~\cite{li2024sscbench}. Unsupervised methods are trained with video sequences and the corresponding ground-truth camera poses from the KITTI-360 dataset. All images are resized to a resolution of $192 \times 640$ pixels. Following \cite{wimbauer2023behind}, we split the dataset into a training set of 98,008 images, a validation set of 11,451 images, and a test set of 446 images. We train the network with both the stereo perspective-camera video sequence and the left and right fisheye-camera video sequences from the KITTI-360 dataset~\cite{Liao2022PAMI}.
For each fisheye image, we follow the resampling process described in \cite{wimbauer2023behind} to obtain the corresponding image using a virtual perspective camera.
To enlarge the overlap among camera frustums, we offset fisheye-camera sequences by ten timestamps relative to the stereo perspective-camera sequence, thereby enhancing the ratio of valid color samples.

\subsubsection{Hyperparameters and Training Configurations}
We train our network~\cite{He_2016_resnet} for 25 epochs using the Adam~\cite{Kingma2014AdamAM} optimizer on an NVIDIA RTX 4090 GPU, with the initial learning rate set to $2\times 10^{-4}$, which is decayed by a factor of 2 during the final 10 epochs.
The batch size is set to 16.
We set the loss weights, $\lambda_r$ and $\lambda_p$, to $1$ and $1 \times 10^{-3}$, respectively.
Furthermore, since the occlusion-aware occupancy polarization mechanism directly utilizes color similarity as a prior to infer the geometric structure between neighboring sampled points, it is reasonable to omit color augmentation for training.
Additionally, we employ horizontal flip augmentation by randomly flipping the input image prior to feeding it into the network. To preserve the original geometric structure of the scene, a corresponding inverse flip is applied to the resulting feature maps before they undergo further processing.


\subsection{Comparisons with State-of-The-Art Methods}

\begin{table}[t!]
	\caption{Quantitative comparison among unsupervised methods on the KITTI-360 dataset.}
	\label{tb:unsup}
	\centering
  \settablefont
	\begin{tabular}{l|cccccc}
		\toprule[1pt]
		Method &$\text{O}_\text{Acc}$ &$\text{O}_\text{Pre}$ &$\text{O}_\text{Rec}$ &$\text{IE}_\text{Acc}$ &$\text{IE}_\text{Pre}$ &$\text{IE}_\text{Rec}$ \\
        \hline
        BTS~\cite{wimbauer2023behind}  & 0.870 & 0.733 & 0.745 & 0.727 & 0.466 & 0.658 \\
        KDBTS~\cite{Han_2024_CVPR}     & 0.871 & 0.746 & 0.731 & 0.722 & 0.463 & 0.682 \\
        KYN~\cite{li2024know}          & 0.861 & 0.746 & 0.654 & 0.671 & 0.402 & \textbf{0.707} \\
        ViPOcc~\cite{feng2025vipocc}   & 0.875 & 0.748 & 0.746 & 0.728 & 0.467 & 0.668 \\
        \hline
        \textbf{Ours}                      & \textbf{0.883} & \textbf{0.763} & \textbf{0.757} & \textbf{0.741} & \textbf{0.475} & 0.676 \\
		\bottomrule[1pt]
	\end{tabular}
\end{table}

\begin{table}[t!]
	\caption{Quantitative comparison with both supervised (denoted as ``S'') and unsupervised (denoted as ``U'') methods on the KITTI-360 dataset. 
    The best results are shown in bold type, with the best unsupervised ones underlined.
    }
	\label{tb:super}
	\centering
	\settablefont
	\begin{tabular}{ll|ccc}
		\toprule[1pt]
		Type & Method 
		&$\text{IoU (\%)}$ 
		&$\text{Pre (\%)}$
		&$\text{Rec (\%)}$
        \\
        \hline
        \multirow{5}{*}{S} 
        & MonoScene     & 37.9 & 56.7 & 53.3 \\
        & TPVFormer     & 40.2 & 59.3 & 55.5 \\
        & VoxFormer     & 38.8 & 58.5 & 53.4 \\
        & OccFormer     & 40.3 & 59.7 & 55.3 \\
        & Symphonies    & 44.1 & \textbf{69.2} & 54.9 \\
        \hline
        \multirow{4}{*}{U} 
        & KDBTS           & 44.6 & 52.7 & 74.3 \\
        & KYN             & 44.4 & \underline{54.0} & 71.4 \\
        & ViPOcc          & 43.1 & 47.2 & \textbf{\underline{83.4}} \\
        & \textbf{Ours}   & \textbf{\underline{45.5}} & 50.8 & 81.4 \\
		\bottomrule[1pt]
	\end{tabular}
\end{table}

\begin{figure*}[!t]
    \centering
    \includegraphics[width=\linewidth]{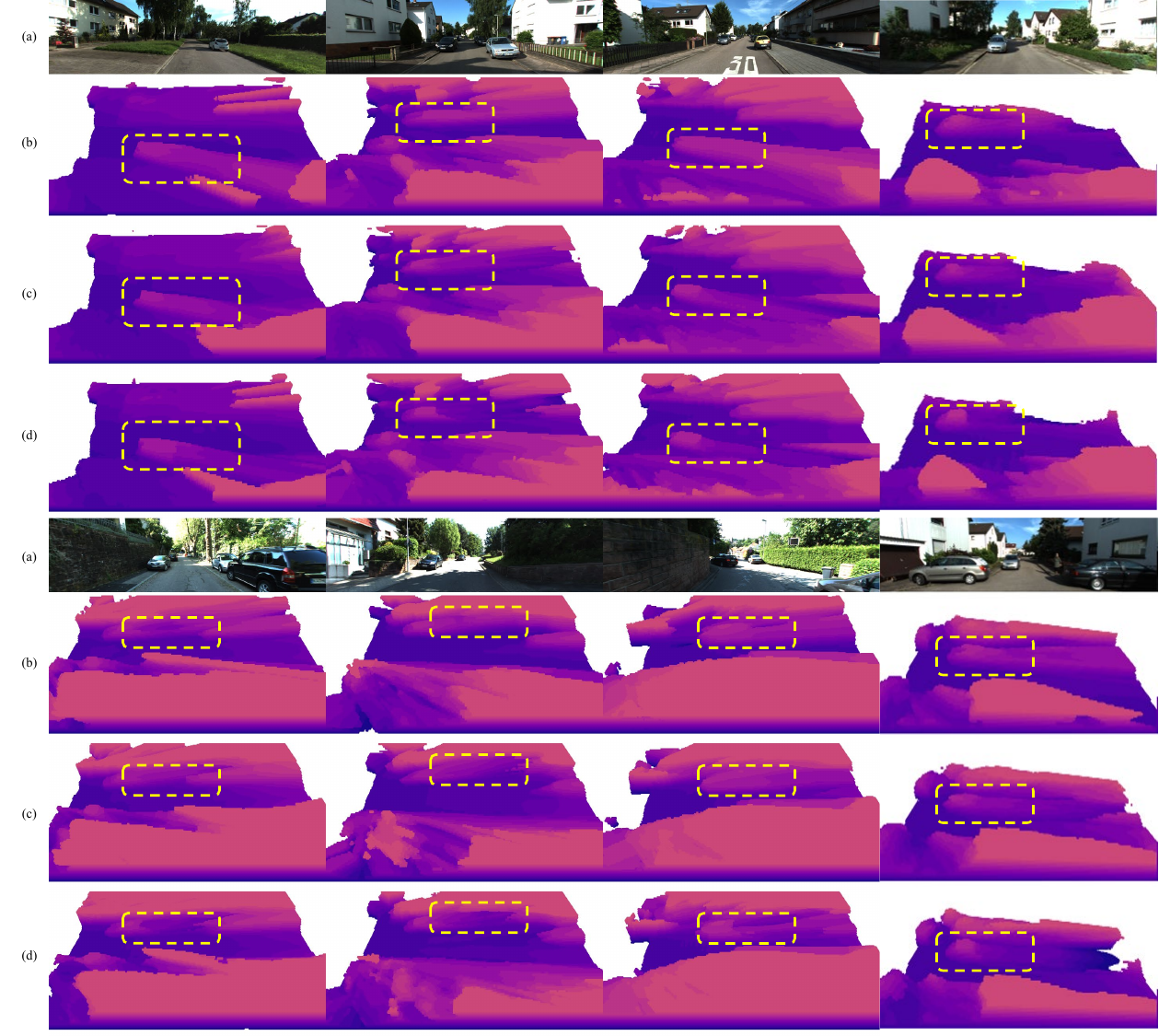}
    \caption{Qualitative comparisons of 3D occupancy prediction on the KITTI-360 dataset: (a) input RGB images; (b) BTS results; (c) ViPOcc results; (d) our results.}
    \label{fig:visualization}
\end{figure*}

\subsubsection{Comparisons with Unsupervised Methods}
We compare our approach with representative unsupervised SoTA methods. As shown in Table \ref{tb:unsup}, our method achieves SoTA performance across the majority of evaluation metrics. In particular, it improves $\mathrm{O}_\text{Acc}$, $\mathrm{O}_\text{Pre}$, $\mathrm{O}_\text{Rec}$, $\mathrm{IE}_\text{Acc}$, and $\mathrm{IE}_\text{Pre}$ by 0.9\%, 2.0\%, 1.5\%, 1.8\%, and 1.7\%, respectively. It is worth noting that although KYN achieves the highest $\mathrm{IE}_\text{Rec}$ score, it incorporates a computationally intensive visual-language network~\cite{li2022language}, which significantly compromises inference efficiency.

Qualitative comparisons are presented in Fig.~\ref{fig:visualization}, where the predicted occupancy grids are visualized from the right side of the scene. Compared to BTS and ViPOcc, our method achieves superior 3D geometry reconstruction, and effectively mitigates trailing effects. The visualization results demonstrate the effectiveness of our proposed polarization mechanism in reasoning occluded occupancy.

\subsubsection{Comparisons with Supervised Methods}
Additionally, our reformulated benchmark suite is aligned with the evaluation protocols used by supervised methods, thereby enabling direct comparison with representative supervised approaches.
The quantitative experimental results presented in Table \ref{tb:super} reveal several noteworthy findings.
Most notably and somewhat unexpectedly, NeRF-based methods achieve IoU scores that are comparable to, or even exceed those of recent supervised approaches, while outperforming most earlier ones. 
We attribute this phenomenon to the limited quality of existing 3D occupancy ground truth, which may introduce misleading supervisory signals and thus hinder the effectiveness of supervised training. In contrast, NeRF-based methods are unaffected by this issue, as they do not rely on such supervision. Moreover, unsupervised methods typically exhibit higher recall than precision, indicating a tendency to overestimate occupied space, particularly in occluded regions where supervisory signals are absent.
In contrast, benefiting from direct supervision in these areas, supervised methods often achieve more balanced metrics. While our method mitigates this imbalance compared to the unsupervised baseline ViPOcc, further improvements are necessary, especially in handling occlusions, which remain a key challenge for unsupervised 3D occupancy prediction.

\begin{table}[t!]
\caption{
  Zero-shot 3D occupancy prediction metrics on the SemanticKITTI dataset.
}
\label{tb:sk-zero-shot}
\centering
\settablefont
\begin{tabular}{l|ccc}
	\toprule[1pt]
	  Method & $\text{IoU (\%)}$ &$\text{Pre (\%)}$ &$\text{Rec (\%)}$    \\
    \hline
    SceneRF & 13.8 & 17.3 & 41.0 \\
    SelfOcc(BEV) & 21.0 & \textbf{37.3} & 32.4 \\
    SelfOcc(TPV) & 22.0 & 34.8 & 37.3 \\
    ViPOcc & 23.6 & 26.9 & 66.8 \\
    Ours & \textbf{24.1} & 27.0 & \textbf{68.7} \\
	\bottomrule[1pt]
\end{tabular}
\end{table}

\subsubsection{Zero-shot Experiments}
To further evaluate the generalizability of the proposed method, we conduct a zero-shot test on the SemanticKITTI dataset~\cite{behley2019iccv} using the pre-trained weights obtained from the KITTI-360 dataset. As presented in Table~\ref{tb:sk-zero-shot}, the proposed method outperforms other SoTA methods including SceneRF~\cite{cao2023scenerf} and SelfOcc~\cite{huang2024selfocc}, demonstrating its exceptional generalizability.

\subsection{Ablation Studies}

\subsubsection{Occupancy Probability Interpretation}
To demonstrate the rationality of our interpreted occupancy probability representation, we evaluate existing unsupervised methods under both the conventional and the proposed representations. Specifically, we utilize pretrained weights from prior works without any modification and exclusively adjust the occupancy probability representation for performance evaluation. The network's performance is then evaluated using the above-mentioned metrics under both interpretations for comparation.

As shown in Table \ref{tb:abl-interpretation}, directly applying our interpreted occupancy probability defined in Eq.~\ref{eq:new}, without any retraining, consistently leads to higher $\mathrm{O}_\text{Acc}$ and $\mathrm{IE}_\text{Rec}$ scores, compared to the conventional representation defined in Eq.~\ref{eq:old}.
This improvement demonstrates that our proposed formulation ensures greater consistency between the training and evaluation protocols and is more suitable for the quantitative evaluation of unsupervised methods.
Furthermore, by leveraging the proposed occupancy probability representation, we observe opposing trends in $\mathrm{IE}_\text{Rec}$ and $\mathrm{IE}_\text{Acc}$, which quantify performance within invisible regions, revealing that existing methods generally misclassify free space as occupied when explicit supervisory signals are absent. This observation corroborates the limitation in current approaches when inferring in occluded regions.

\begin{table}[t!]
	\caption{Ablation study on the occupancy probability interpretation. The conventional occupancy probability interpretation is given in \eqref{eq:old}, whereas ours is given in \eqref{eq:new}.}
	\label{tb:abl-interpretation}
	\centering
	\settablefont
	\begin{tabular}{lc|ccc}
		\toprule[1pt]
		  Method & Representation
		&$\text{O}_\text{Acc}$ 
		&$\text{IE}_\text{Acc}$
		&$\text{IE}_\text{Rec}$
        \\
        \hline
        \multirow{2}{*}{BTS} 
        & Conventional     & 0.867 & \textbf{0.756} & 0.606 \\
        & Ours     & \textbf{0.870} & 0.727 & \textbf{0.658} \\
        \hline
        \multirow{2}{*}{KDBTS} 
        & Conventional     & 0.868 & \textbf{0.750} & 0.618 \\
        & Ours     & \textbf{0.871} & 0.722 & \textbf{0.682} \\
        \hline
        \multirow{2}{*}{ViPOcc} 
        & Conventional     & 0.873 & \textbf{0.757} & 0.608 \\
        & Ours     & \textbf{0.875} & 0.728 & \textbf{0.668} \\
		\bottomrule[1pt]
	\end{tabular}
\end{table}

\begin{figure}[!t]
    \centering
    \includegraphics[width=\linewidth]{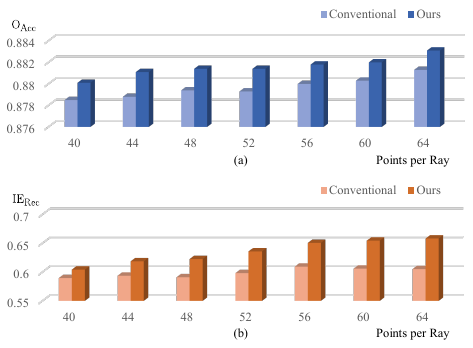}
    \caption{Comparison between the conventional and proposed occupancy probability representations in terms of (a) $\text{O}_\text{Acc}$ and (b) $\text{IE}_\text{Rec}$ across varying numbers of sampled points per ray.}
    \label{fig:abl-inter-varp}
\end{figure}

To further demonstrate the robustness of the proposed occupancy probability representation under varying point sampling intervals along rays, we train the baseline network~\cite{feng2025vipocc} with different numbers of sampled points per ray, while maintaining fixed near and far bounds, as defined in the NeRF framework. 
As illustrated in Fig.~\ref{fig:abl-inter-varp}, unlike the conventional representation, which suffers from fluctuations in network output magnitude under varying sampling densities, our representation maintains consistent performance, demonstrating greater stability to changes in point sampling strategies, as discussed above.

\begin{table}[t!]
	\caption{Ablation study on the occlusion-aware occupancy polarization mechanism.}
	\label{tb:abl-polarization}
	\centering
	\settablefont
	\begin{tabular}{lc|ccc}
		\toprule[1pt]
		  Baseline & $\mathcal{L}_p$ 
		&$\text{O}_\text{Acc}$ 
		&$\text{IE}_\text{Acc}$
		&$\text{IE}_\text{Rec}$
        \\
        \hline
        \multirow{2}{*}{BTS}
        & \ding{55}     & 0.870 & 0.727 & 0.658 \\
        & \checkmark    & \textbf{0.880} & \textbf{0.737} & \textbf{0.667} \\
        \hline
        \multirow{2}{*}{KDBTS} 
        & \ding{55}     & 0.871 & 0.722 & \textbf{0.682} \\
        & \checkmark    & \textbf{0.879} & \textbf{0.725} & \textbf{0.682} \\
        \hline
        \multirow{2}{*}{ViPOcc} 
        & \ding{55}     & 0.875 & 0.728 & 0.668 \\
        & \checkmark    & \textbf{0.883} & \textbf{0.741} & \textbf{0.676} \\
		\bottomrule[1pt]
	\end{tabular}
\end{table}

\subsubsection{Occlusion-Aware Occupancy Polarization.}
To validate the effectiveness of the occlusion-aware occupancy polarization mechanism, we incorporate its corresponding loss $\mathcal{L}_{p}$ into the overall loss function and retrain several baseline networks for comprehensive comparisons.
As shown in Table~\ref{tb:abl-polarization}, the mechanism consistently improves all evaluation metrics across all baseline networks, with maximum improvements of 1.1\%, 1.8\% and 1.4\% on $\mathrm{O}_{\text{Acc}}$, $\mathrm{IE}_{\text{Acc}}$ and $\mathrm{IE}_{\text{Rec}}$, demonstrating its general efficacy.

When designing the polarization mechanism, our goal is to establish an indicator that reflects whether two adjacent 3D points along a ray correspond to the same object when projected into a given source view.
Other than relying on the color difference, we exploit the difference in pseudo depth predicted by a vision foundation model~\cite{yang2024depth}. Specifically, for each pair of adjacent samples along a ray, we obtain their projections in the source view and examine the discrepancy of their image-level signals (\textit{e.g.}, color or pseudo depth) as an indicator to determine whether these projections lie on the same object.

\begin{table}[t!]
	\caption{Ablation study on the signals of occlusion-aware occupancy polarization mechanism.}
	\label{tb:abl-polarization-pd}
	\centering
	\settablefont
	\begin{tabular}{lc|cccccc}
		\toprule[1pt]
		  Baseline & Signal for building $\mathcal{L}_p$ 
		&$\text{O}_\text{Acc}$ 
		&$\text{IE}_\text{Acc}$
		&$\text{IE}_\text{Rec}$
        \\
        \hline
        \multirow{2}{*}{BTS}
        & Pseudo depth     & 0.879 & 0.734 & 0.644 \\
        & RGB intensity   & \textbf{0.880} & \textbf{0.737} & \textbf{0.667} \\
        \hline
        \multirow{2}{*}{KDBTS} 
        & Pseudo depth     & 0.878 & 0.724 & 0.679 \\
        & RGB intensity    & \textbf{0.879} & \textbf{0.725} & \textbf{0.682} \\
        \hline
        \multirow{2}{*}{ViPOcc} 
        & Pseudo depth     & 0.881 & 0.740 & 0.655 \\
        & RGB intensity   & \textbf{0.883} & \textbf{0.741} & \textbf{0.676} \\
		\bottomrule[1pt]
	\end{tabular}
\end{table}

Experimental results presented in Table \ref{tb:abl-polarization-pd} suggest that RGB-based indicators consistently outperform pseudo-depth-based ones across baseline models. This finding is somewhat unexpected, given that depth maps typically provide more reliable geometric cues and are generally more robust to texture ambiguity and color similarity. 

We hypothesize that this phenomenon stems from the distinct signal transition characteristics across object boundaries.
For two adjacent points located on different objects, the pseudo-depth differences can vary significantly depending on the geometric structure of the scene. In particular, when these points lie on different objects but have similar depth values, their pseudo-depth values are often close, making it difficult to distinguish inter-object cases from intra-object ones.
In contrast, due to variations in lighting, material, and texture across different objects, RGB intensities tend to differ markedly across object boundaries, even when depth values are similar.
This observation suggests that RGB intensity is generally more reliable than pseudo-depth for inferring object-level consistency.

We further conduct a hyperparameter sensitivity analysis. Since $\lambda_r$ acts as the primary loss weight and is conventionally fixed at 1 without compromising training stability, we vary only $\lambda_p$ to examine its impact on occlusion-region accuracy, as reported in Table~\ref{tb:lambda-p}. Specifically, $\lambda_p$ is varied from $10^{-4}$ to $10^{-2}$. The model achieves its best performance at $\lambda_p = 1 \times 10^{-3}$. Moreover, the variations in $\mathrm{O}_{\text{Acc}}$, $\mathrm{IE}_{\text{Acc}}$, and $\mathrm{IE}_{\text{Rec}}$ across different $\lambda_p$ settings are limited to within 0.5\%, 0.9\%, and 3.6\%, respectively, indicating that the proposed method is robust to the choice of $\lambda_p$.


\begin{table}[t!]
\caption{
Sensitivity analysis of $\lambda_p$.
}
\label{tb:lambda-p}
\centering
\settablefont
\begin{tabular}{l|ccc}
\toprule[1pt]
$\lambda_p$ & $\mathrm{O}_{\text{Acc}}$ & $\mathrm{IE}_{\text{Acc}}$ & $\mathrm{IE}_{\text{Rec}}$ \\
\hline
$1 \times 10^{-4}$ & 0.880 & 0.740 & 0.654 \\
$5 \times 10^{-4}$ & 0.879 & 0.737 & 0.661 \\
$1 \times 10^{-3}$ & \textbf{0.883} & \textbf{0.741} & \textbf{0.676} \\
$5 \times 10^{-3}$ & 0.880 & 0.737 & 0.654 \\
$1 \times 10^{-2}$ & 0.878 & 0.734 & 0.651 \\
\bottomrule[1pt]
\end{tabular}
\end{table}

\section{Discussion}
\label{sec:discussion}

\begin{figure}[!t]
    \centering
    \includegraphics[width=\linewidth]{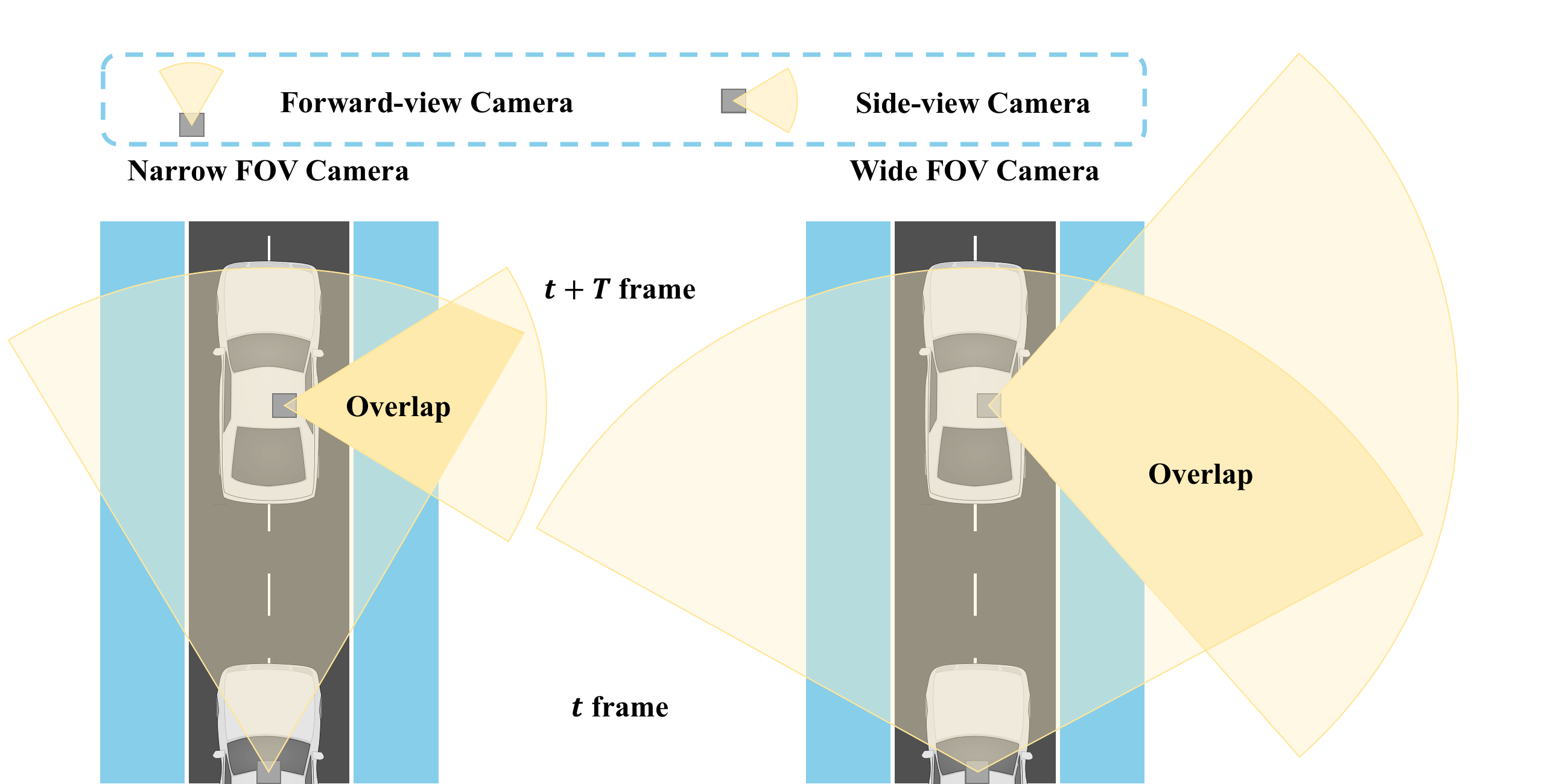}
    \caption{An illustration of the limitation related to camera field of view.}
    \label{fig:failurecases-sup}
\end{figure}

Despite achieving SoTA performance, the proposed method still presents a known limitation which emerges when we conduct experiments on nuScenes datasets~\cite{caesar2020nuscenes}. To be specific, in each training iteration we use the front-camera image at frame $t$ together with the back-left, back-right camera images at frame $t+6$. The offset of 6 frames is an empirical choice to maximize the spatial overlap between views. However, current unsupervised 3D occupancy prediction methods (BTS, KDBTS, KYN, ViPOcc and Ours) struggle to converge.

We attribute this phenomenon to the narrow camera field of view in this datasets. During training, forward-view cameras at frame $t$ and side-view cameras at frame $t+T$ are selected (where $T$ is a fixed parameter).
Current unsupervised 3D occupancy prediction methods fundamentally rely on multi-view consistency constraints. As discussed in Sec.~\ref{sec:op}, all supervisory signals are derived from multi-view 2D images and supervise the network’s occupancy predictions only within regions where camera views overlap. Consequently, the volume of these overlapping regions, which is positively correlated with the cameras' horizontal field of view (FOV), directly determines the effectiveness of the training process.
However, compared with the KITTI-360 dataset (FOV = $103^\circ$), nuScenes (FOV = $64^\circ$) dataset provides much narrower camera views. As illustrated in Fig.~\ref{fig:failurecases-sup}, the resulting low proportion of overlapping regions severely hinders the establishment of multi-view supervisory signals, preventing the metrics from converging.
To further validate this phenomenon, we crop the image in KITTI-360 dataset to reduce the FOV to $64^\circ$, matching that of nuScenes, and observe the same issue: the network’s performance metrics cease to converge. Our future work will investigate improved dataset organization strategies to increase the volume of overlapping regions.

\section{Conclusion and Future Work}
\label{sec:conclusion}

In this paper, we first addressed a critical limitation in the existing unsupervised monocular 3D occupancy prediction benchmark: the spatial inconsistency between training and evaluation protocols. To this end, we developed an interpretable, opacity-based representation of occupancy probability and introduced a coordinate-transformed sampling algorithm for voxel-wise occupancy prediction, contributing a consistent and reliable evaluation protocol aligned with those used by supervised methods. In addition, to compensate for the inherent lack of photometric supervision revealed by the proposed benchmark, we leveraged multi-view visual cues and introduced an occlusion-aware occupancy polarization mechanism, which proves to be compatible across all baseline networks. Extensive experiments conducted with both supervised and unsupervised methods on the reformulated benchmark validate the rationality of our interpreted occupancy probability, the alignment between training and evaluation protocols, and the effectiveness of the proposed occlusion-aware occupancy polarization mechanism.
In the future, we plan to extend this benchmark with instance-level evaluation metrics to more comprehensively evaluate the model’s capabilities.

\bibliographystyle{IEEEtran}
\bibliography{ref}

\end{document}

%% file: packages.tex
\usepackage{ulem}
\usepackage{url}

\usepackage{scalerel}
\usepackage{color}
\usepackage{tikz}
\usetikzlibrary{svg.path}
\definecolor{orcidlogocol}{HTML}{A6CE39}
\tikzset{
  orcidlogo/.pic={
    \fill[orcidlogocol] svg{M256,128c0,70.7-57.3,128-128,128C57.3,256,0,198.7,0,128C0,57.3,57.3,0,128,0C198.7,0,256,57.3,256,128z};
    \fill[white] svg{M86.3,186.2H70.9V79.1h15.4v48.4V186.2z}
                 svg{M108.9,79.1h41.6c39.6,0,57,28.3,57,53.6c0,27.5-21.5,53.6-56.8,53.6h-41.8V79.1z M124.3,172.4h24.5c34.9,0,42.9-26.5,42.9-39.7c0-21.5-13.7-39.7-43.7-39.7h-23.7V172.4z}
                 svg{M88.7,56.8c0,5.5-4.5,10.1-10.1,10.1c-5.6,0-10.1-4.6-10.1-10.1c0-5.6,4.5-10.1,10.1-10.1C84.2,46.7,88.7,51.3,88.7,56.8z};
  }
}
\newcommand\orcidicon[1]{\href{https://orcid.org/#1}{\mbox{\scalerel*{
\begin{tikzpicture}[yscale=-1,transform shape]
\pic{orcidlogo};
\end{tikzpicture}
}{|}}}}
\usepackage{hyperref}
\hypersetup{
    colorlinks=true,
    linkcolor=blue,
    filecolor=black,      
    urlcolor=magenta,
    citecolor=purple
}
\usepackage{tikz-network}